\documentclass[letterpaper, 10 pt, conference]{ieeeconf}  % Comment this line out if you need a4paper

\IEEEoverridecommandlockouts                              % This command is only needed if 
\usepackage{graphics} % for pdf, bitmapped graphics files
\usepackage{epsfig} % for postscript graphics files
\usepackage{mathptmx} % assumes new font selection scheme installed
\usepackage{times} % assumes new font selection scheme installed
\usepackage{amsmath} % assumes amsmath package installed
\usepackage{multirow}
\usepackage{amssymb}  % assumes amsmath package installed
\usepackage{color}
\usepackage{gensymb}
\usepackage{siunitx}
\usepackage{lipsum}
\usepackage{mathtools, nccmath}
\usepackage{cuted}

\usepackage{hyperref}
\hypersetup{
    colorlinks=true,
    linkcolor=blue,
    filecolor=magenta,      
    urlcolor=blue,
}

\newcommand{\myparagraph}[1]{\vspace{0.05in}\noindent\textbf{#1}}

\newcommand{\mysubscript}[1]{\raisebox{0.5ex}{\scriptsize#1}}

\title{\LARGE \bf
Tactile-RL for Insertion:\\ Generalization to Objects of Unknown Geometry
}

\author{
  \authorblockN{Siyuan Dong$^{1}$, 
Devesh K. Jha$^{2}$, Diego Romeres$^{2}$, Sangwoon Kim$^{1}$, Daniel Nikovski$^{2}$ and Alberto Rodriguez$^{1}$} 
  \authorblockA{
     $^{1}$Massachusetts Institute of Technology, $^{2}$Mitsubishi Electric Research Laboratories\\
    {\tt\small $^{1}$<sydong,sangwoon,albertor>@mit.edu, $^{2}$<jha,romeres,nikovski>@merl.com}} \
\thanks{This research is supported by Mitsubishi
Electric Research Laboratories (MERL), and the HKSAR Innovation and Technology Fund (ITF) ITS-104-19F. MERL researchers were not supported by HKSAR.}
}

\begin{document}

\maketitle

\thispagestyle{empty}
\pagestyle{empty}

% \begin{strip}
% \centering
% \vspace{-15mm}
% \includegraphics[width=\textwidth]{figure/teaser_v4.png}
% Insertion task with the proposed Tactile RL policy.
% \label{fig:tactile-RL-abstract}
% \end{strip}

%%%%%%%%%%%%%%%%%%%%%%%%%%%%%%%%%%%%%%%%%%%%%%%%%%%%%%%%%%%%%%%%%%%%%%%%%%%%%%%%
\begin{abstract}

Object insertion is a classic contact-rich manipulation task. The task remains challenging, especially when considering general objects of unknown geometry, which significantly limits the ability to understand the contact configuration between the object and the environment.
We study the problem of aligning the object and environment
%implicitly learn to localize external contacts
with a tactile-based feedback insertion policy.
The insertion process is modeled as an episodic policy that iterates between insertion attempts followed by pose corrections. We explore different mechanisms to learn such a policy based on Reinforcement Learning.
The key contribution of this paper is to demonstrate that it is possible to learn a tactile insertion policy that generalizes across different object geometries, and an ablation study of the key design choices for the learning agent: 1) the type of learning scheme: supervised vs. reinforcement learning; 2) the type of learning schedule: unguided vs. curriculum learning %through a series of increasingly complex assembly scenarios
; 3) the type of sensing modality: force/torque vs. tactile; and 4) the type of tactile representation: tactile RGB vs. tactile flow.
We show that the optimal configuration of the learning agent (RL + curriculum + tactile flow) exposed to $4$ training objects yields an closed-loop insertion policy that inserts $4$ novel %everyday 
objects with over $85.0\%$ success rate and within $3\sim4$ consecutive attempts. 
% The learning is done in a real robotic system within 500 training episodes and within 8 hours of robot time. 
Comparisons between F/T and tactile sensing, shows that while an F/T-based policy learns more efficiently, a tactile-based policy provides better generalization. See supplementary video and results at \href{https://sites.google.com/view/tactileinsertion}{https://sites.google.com/view/tactileinsertion}. 
%
%This RL policy outperforms all baseline policies in success rate, number of insertion attempts, and speed of convergence.
\end{abstract}

%%%%%%%%%%%%%%%%%%%%%%%%%%%%%%%%%%%%%%%%%%%%%%%%%%%%%%%%%%%%%%%%%%%%%%%%%%%%%%%%
\section{Introduction}\label{sec:introduction}
Localizing contacts between objects and their environment is central to many contact-rich robotic manipulation tasks~\cite{mason2018toward}. 
Consider for example the classical tasks of inserting a peg in a hole or packing an object in a box. A mismatch between the real and modeled object's shape or between the real and target object's pose, generates unexpected contacts between the object and the hole, potentially leading to complex contact interlocking configurations~\cite{yu2018realtime}. 

The location of contacts during insertion depend on the geometrical attributes of the object and hole.
%(i.e., shape, size, etc.). 
%
Techniques to explicitly recover the location of those contacts, usually require knowledge of object geometries and/or are only applicable to limited types of contact configurations~\cite{Bicchi1993,Li2015,Manuelli2016,yu2018realtime}. The estimation of those contacts without knowledge of object geometry is also possible in some cases by aggregating tactile information over time~\cite{ma2021extrinsic}.

\begin{figure}[t]
	\centering
	\includegraphics[width=\linewidth]{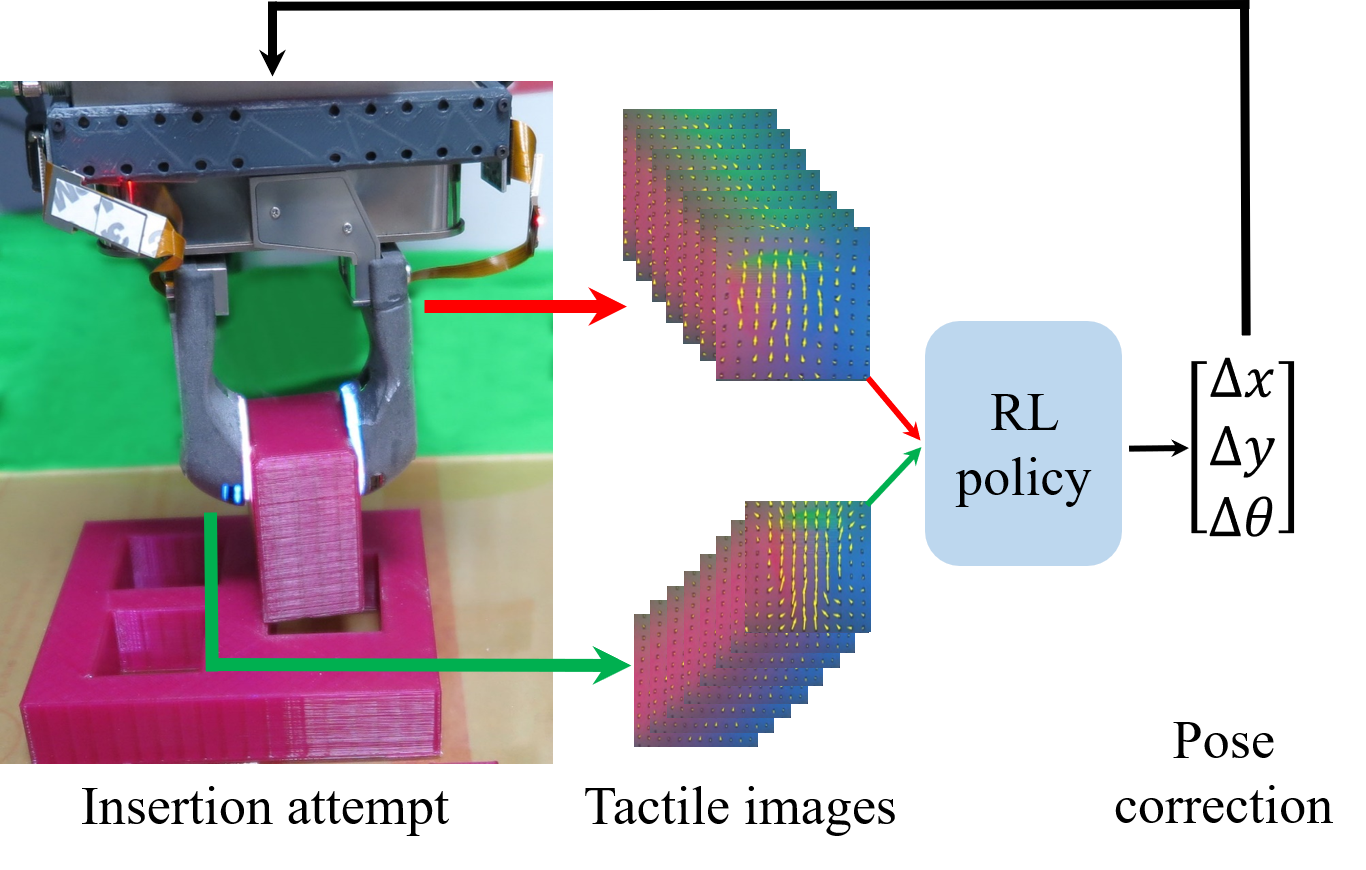}
	\vspace{-20pt}
	\caption{Insertion task with the proposed Tactile RL policy.}
	\label{fig:tactile-RL-abstract}
	\vspace{-17pt}
\end{figure}

In this paper, alternatively, we are interested in feedback-based mechanisms to interactively correct the miss-alignment between object and hole without making use of the geometry of the shapes that generated it and without making explicit inference of the contact location. 
% In the absence of a geometric model, the robot has to rely on available sensors to build a feedback-based policy that indirectly estimates and corrects for the miss-alignment between object and environment.
Tactile sensors co-located at the gripper fingers (see Figure~\ref{fig:tactile-RL-abstract}) are better positioned to capture contact events than vision sensors, which suffer from occlusions and limited accuracy. Force/Torque (F/T) sensors can also be co-located at fingers or wrists. 
In this work, we investigate the use of F/T and high-resolution tactile sensors to guide peg-in-hole insertion policies, and to generalize to different objects while accounting for variations in contact formations, in grasps, and in environments, i.e., a general \emph{tactile insertion agent}.

The key challenge to a general insertion policy is that, in the absence of geometric models, the contact state is not fully observable from a single time instant, neither for force or tactile sensing. The insertion agent, then, needs to consider policies beyond greedily trying to estimate and correct the miss-alignment between part and hole, as was shown in previous schemes based on self-supervised learning, which showed limited generalization to objects and contact configurations~\cite{dong2019tactile, yu2018realtime}
%
%An important point to note is that, even for tactile sensors co-located at the fingers, the contact state between the object and environment, is not fully observable from a single time instant. The insertion agent needs to explore and reason about correlations between attempts to successfully recover and correct misalignments. 
%
%This is a key limitation in previous works based on learning self-supervised insertion agents with limited generalization to objects and contact configurations ~\cite{dong2019tactile, yu2018realtime}.
%
In recent work~\cite{dong2019tactile}, we demonstrated that high-resolution tactile sensors such as GelSlim~\cite{taylor2021iros} allow 
%the insertion agent to extract a interpret geometry and force information of the contacts with the environment, and enable 
feedback correction for novel objects. However, the simplified insertion environment (a slot with constrain in one axis) only covers a very small set of possible contact formations in a object insertion problem. 

In this paper, we explore the more complex and general problem of \emph{inserting multiple objects into multiple environments with a single policy}, and we ablate several important design choices for an RL tactile-based insertion agent. In particular we evaluate:
\begin{itemize}
    \item \textbf{Supervised vs. Reinforced} We study the importance of the sequential nature of the learning process in rewarding successful insertions, and learn an episodic RL agent that iterates between insertion attempts and estimating alignment corrections (see Figure~\ref{fig:tactile-RL-abstract}). While the supervised agent suffers in the more constrained insertion environments, the RL agent shows significantly better performance. The RL formulation, with the ability to consider delayed rewards, opens the possibility to search through non-greedy policies.
    \item \textbf{Curriculum training.} We show that a modest guidance of the learned policy to progressively tackle more complex environments (wall $\rightarrow$ corner $\rightarrow$ U $\rightarrow$ hole) is effective at increasing the data efficiency .
    \item \textbf{Tactile representation.} We show that learning an insertion agent from tactile flow (represented with marker array) yields much better generalization to variations in object and grasp geometry than directly learning from raw tactile RGB images. Tactile flow is effective at preventing overfitting to object surface texture.

\end{itemize}

%\sdnote{add the conclusion from the comparison with F/T sensor}
Since the dynamics of insertion and the dynamics of tactile sensing are both guided by complex and difficult to simulate contact mechanics, the use of real experiments and the need for data-efficiency are key drivers of this work. 
We also compare our tactile-RL agent with a policy trained using F/T sensors and show that the tactile-RL agent provides better generalization to objects of different geometry. These experiments also demonstrate the limitations of F/T measurements that can be alleviated using tactile sensors.
All the experiments in this paper are done in a real system, with real data, and in real time. The insertion RL agent is learned in 500 episodes (8 hours of robot time). For this purpose, we developed an experimental setup (Fig.~\ref{fig:setup_NN}) that allows to run tactile-based automated insertion experiments under controlled variations on the initial grasp on different objects, and with an automatic resetting mechanism based on object-specific alignment features.

\section{Related Work}\label{sec:related_work}
% \myparagraph{Peg Insertion} 
Peg insertion tasks have been studied for a long time, due to their importance in manufacturing. Early work on peg-in-hole insertion algorithms were based on developing passive compliance devices, \cite{10.1115/1.3149634,Whitney_Nevins77}. These devices, as well as similarly operating linear impedance controllers, can successfully complete insertions  with chamfered holes, as long as the initial contact point is on the chamfer. Other methods use model-based approaches to directly estimate the pose of the peg relative to the hole with force feedback~\cite{bruyninckx1995peg, came1989assembly}. However, these require knowledge of object models and/or are constrained to a single object type. Furthermore, there is a fundamental ambiguity in the mapping from alignment errors to sensed forces --- in many contact situations, it is many-to-one, so it cannot be inverted directly \cite{Newman2001InterpretationAssembly}. 

% For example, when a force/torque (F/T) sensor mounted at the wrist is used, in cases when the edge of the peg is completely out of the hole, many relative positions of the peg with respect to the hole would result in the same force readings, so the exact relative position cannot be recovered~\cite{Newman2001InterpretationAssembly,Sharma2013IntelligentStrategies,Inoue2017DeepTasks}.  %For such cases, a suitable search motion needs to be performed until at least part of the peg is in the hole. Blind search motions scan the surface in a grid-like or circular motion and tend to be slow and unreliable. A much better approach is to use a search strategy that is informed by multiple F/T readings in order to estimate the misalignment and find the hole \cite{Newman2001InterpretationAssembly,Sharma2013IntelligentStrategies,Inoue2017DeepTasks}. Most such methods use different controllers for the search phase and for the actual insertion phase.  

An early RL approach to the actual insertion phase was proposed in \cite{gullapalli1993learning}, similarly based on an episodic RL setting from observations of position and forces from an F/T sensor. Since then, several RL approaches have been proposed, mainly based on sensing and acting on the positional state and force-torques of the robotic manipulators. In \cite{DBLP:journals/corr/LevineWA15}, time-varying linear-Gaussian controllers combined together via a guided policy search component are presented. 
% The authors obtained good generalization and data efficiency performance on a variety of insertion tasks. 
%
\cite{luo2018deep} studies the insertion of a rigid peg into a deformable hole, where the position of the hole relative to the robot is known, possibly with minor uncertainty. This paper combines robot-state information and filtered force/torque readings as the input of the controller. 
Vision-based RL approaches have also been studied. For example, \cite{schoettler2019deep} proposes using residual RL, to correct of a nominal policy produced by state of the art off-policy algorithms to perform connector insertions.
Finally, we also find multi-modal sensing approaches. In~\cite{lee2018making} authors study the importance of each modality (vision, F/T and positional sensing) for peg insertion tasks in simulation which then transfer the policy into the real system. The authors of \cite{liu2020understanding} proposed empirical studies on the effects of the three different modalities for insertion tasks in a real robotic set-up using behavioral cloning.
 
Most RL approaches above do not use force/torque or tactile signals to extract geometric information of contacts, but rather for detecting in/out of contact. 
In~\cite{dong2019tactile} the authors demonstrate the potential use of vision-based tactile sensors by explicitly estimating the contact error with supervised learning, and driving the insertion with a proportional controller. In this work, we directly train an RL policy to guide the insertion with tactile feedback from 2 vision-based tactile sensors in a more challenging insertion scenario with different objects and environments. %We show improvement in both the success rate and the average number of attempts. 

% \myparagraph{Curriculum for RL} One of the primary challenges of RL algorithms is their low data efficiency, which limits their application to real robot settings. Curriculum learning~\cite{bengio2009curriculum}, which breaks down the complexity of the problem in a sequence of increasingly complex tasks is a promising solution. RL with curriculum training has been used either explicitly or implicitly in many robotics tasks. Luo~\textit{et al}.~\cite{luo2020accelerating} designed a curriculum in a robotic arm reaching task by increasing the reaching precision gradually. In OpenAI's Rubik’s cube work\cite{akkaya2019solving}, curriculum learning was used in Automatic Domain Randomization (ADR) through an increase in the complexity of the rendered environment. In~\cite{zhang2020automatic}, the authors applied curriculum learning to automatically generating goals that are neither too easy nor too hard for the current agent and showed advantages of the method in 13 multi-goal robotic tasks from the OpenAI Gym. In this paper, we use curriculum training by manually specifying a series of insertion environments and gradually increasing the number of insertion objects. This allows the agent to gradually learn a policy in several different environments, and also helps us gain meaningful insight into the underlying problem. 
% Another relatively data efficient algorithm was proposed in \cite{DBLP:journals/corr/abs-1708-04033}, where the authors develop a teach-less approach for precise peg-in-hole tasks with a discrete action space.

\section{Methodology}
\label{sec:method}
In this section we introduce the method by addressing the following 4 questions: 1) Why is tactile sensing essential for capturing contacts with the environment? 2) Why is RL a good fit for the insertion task? 3) Why is curriculum learning needed in the RL training? 4) How different representations of the tactile signal affect the learning agent?

\subsection{Tactile Sensing for Contact Localization}
Humans routinely do blind insertions by feeling the interaction between object and environment through the grasp on the object. Contact with the environment generates small forces and small object displacements which are captured by force and tactile sensors in the fingers and hands.
Similarly, we use two high-resolution tactile sensors to capture the subtle rotation signals of the object during contact in the insertion task. In~\cite{dong2019tactile}, we have shown that it is possible in some configurations to recover explicit external contacts from those subtle rotation signals. These sensors capture a depth image during contact as well as the in-plane displacement of an array of markers on the sensor surface.

\begin{figure*}[t]
	\centering
	\includegraphics[width=\linewidth]{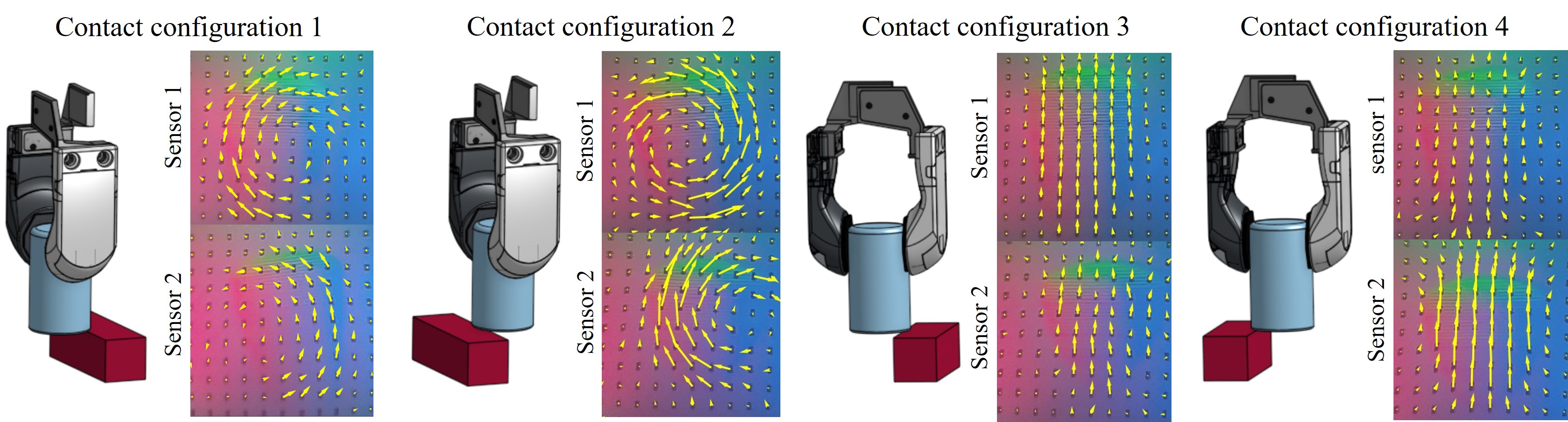}
    \vspace{-15pt}
	\caption{Tactile signals for four contact configurations. In the first 2 cases, the object contacts an edge in the environment that is perpendicular to the plane of the sensors, and consequently the object rotates in the plane of the sensors. Note that the rotation directions are opposite in cases 1 and 2. In the last 2 cases, the object contacts an edge parallel to the plane of the sensor, and consequently rotates out from the sensor plane. Note that the relative vertical displacements in the two sensors is different for cases 3 and 4.}
	\label{fig:tactile}
	\vspace{-10pt}
\end{figure*}

Figure~\ref{fig:tactile} illustrates the signals generated on the sensors under different contact configurations. In the example, the tactile images are captured during the contact phase between a cylindrical object and an edge in the environment, while the gripper moves down vertically. To visualize the flow of the tactile imprint during that phase, we use yellow arrows to draw the relative displacement of the marker field during that contact phase for Sensor1 (black) and Sensor2 (silver). 

The key observation is that some different contact locations generate different patterns of tactile flow. 
In the first two configurations, the object contacts an edge that is perpendicular to the surface of the tactile sensors. In this case the object rotates in the plane of the sensors. The captured patterns of rotation contains information about the external contact point.
Similarly, in the last two configurations, the object contacts an edge in the environment that is parallel to the surface of the tactile sensors. In this case the object rotates out from the plane of the sensors. These capture the vertical displacements of the contacts. 
For simple contact configurations as in Fig.~\ref{fig:tactile}, and if the geometry of the object and environment are known, it is not difficult to directly describe how the tactile flow should look like. In general, however, the relationship becomes more involved for complex contact formations. In any case, the tactile image sequence during the contact period from two tactile sensors contains key information to the location of external contacts. 

\subsection{Deep RL Controller}
%supervised learning doesn't work and why?
In previous work Dong and Rodriguez~\cite{dong2019tactile} used a supervised learning (SL) model to directly estimate the misalignment error between object and hole from GelSlim~\cite{taylor2021iros} tactile feedback in the form of image sequences. While this approach might work well in simple scenarios, due to the partial observity, the SL model may fall apart and make the controller diverge. 
We will show in Sec.~\ref{sec:experiment} that this approach does not work well in more complex environments such as a rectangular hole. 

A more general solution is to model the insertion problem as a sequential decision problem and use a reinforcement learning (RL) algorithm to find the optimal policy. The input to the RL framework is the tactile image sequence during contact. The action is the robot motion in operational space. The reward is related not only to the pose error corrected in each step, but the success or failure of the insertion. The optimal RL policy tries to achieve the maximum reward by inserting the object inside the hole within a sequence of actions, instead of greedily correcting the pose error in each step. In this way, the policy can potentially perform better under some partially observable states, where it is not possible to estimate the full pose error with the tactile signals. This is the key advantage of the RL policy over an SL policy detailed in Section~\ref{sec:experiment}.

\subsection{Curriculum Learning}
% why we need to use Curriculum training 
Due to the challenges of simulating the tactile signal with accuracy during contact, we train the insertion policy on a real robot. In order to improve the data efficiency of RL and make the problem feasible to solve in a reasonable training time, we explore the use of curriculum learning. 

% How to design the curriculum learning 
We design the learning curriculum by scaling the complexity of the insertion environment. In Fig.~\ref{fig:setup_NN}, we introduce 4 insertion environments shown in red: line wall $\rightarrow$ corner wall $\rightarrow$ U wall $\rightarrow$ hole. Note that each environment adds one more constraint to the previous environment, and forces the insertion policy to learn to handle that new constraint. 
Intuitively, the curriculum is asking the RL policy to incrementally learn to: 1) distinguish the contact direction in one axis in the line wall environment; 2) reason about the contact directions in 2 axis in the corner environment; 3) control the step size to avoid overshoot in one axis  and rotation in the U environment; and 4) control step size in both axis and rotation in the hole environment. 

In order to generalize the RL policy to different objects, we train the policy with 4 objects with different shapes (cylinder, hexagonal cylinder, elliptical cylinder, and cuboid) shown in Fig.~\ref{fig:setup_NN}. We also use curriculum learning by gradually increasing the number of objects to insert in the line environment. Generally, the cylinder, hexagonal cylinder, and elliptical cylinder are easier to insert, since they are less sensitive to rotational error. The cuboid is the most difficult to insert because a small misalignment in yaw will block the insertion. %Curriculum training with objects is used only in the first environment, and we train with all objects together in the rest of the environments. 

\subsection{Tactile representation}
% extracting feature is import, 

The choice of the tactile signal representation affects the generalization of the policy to new objects, since the combination of a high-dimensional space, and a small number of training objects can easily lead to overfitting. % Using a good representation of the tactile signal helps the RL policy converge faster and generalize better to new objects.
High-resolution tactile images encode rich information of the contact surface. Some information content such as texture might not be relevant to the insertion task, while some other information might be more closely related to the location of external contacts, such as force distribution and tactile flow.
Here we study the effect of data representation of the tactile signal on the RL policy using 2 different types of data input: 1) the raw RGB image, and 2) the marker flow image.

\section{Experimental Details}
\label{sec:experiment}
\subsection{Experimental Setup}
\begin{figure*}[t]
    \centering
	\includegraphics[width=\linewidth]{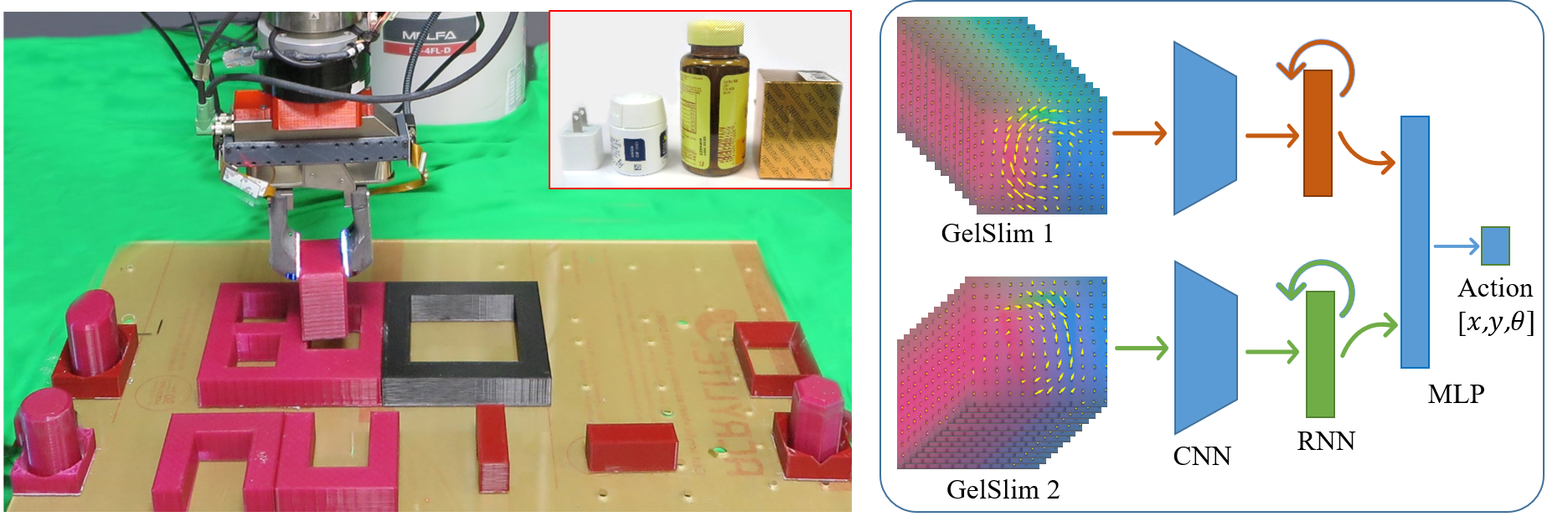}
    \vspace{-15pt}
	\caption{Left: the experimental setup including a Mitsubishi Electric RV-4FL model robot arm, a WSG-32 parallel-jaw gripper, 2 revised GelSlim sensors, 4 sets of insertion environments, 4 training objects (cylinder and elliptical cylinder on the left,  hexagonal cylinder on the right, cuboid held), and 4 novel everyday objects (phone charger, small bottle, big bottle, and box). Right: The Actor Neural Network model in TD3. The CNN includes 4 convolutional layers with 5, 3, 3, 3 kernel size and extracts $1\times512$ features. The RNN are 2 layers of LSTM with 512 memory units and output .$1\times512$ features. The MLP includes 2 fully connected layers with 512 and 256 hidden units. The marker flow is for visualization, not included in the training.}
	\label{fig:setup_NN}
	\vspace{-15pt}
\end{figure*}

Figure~\ref{fig:setup_NN} shows the experimental setup, including a 6-DoF Mitsubishi Electric RV-4FL robot arm, a WSG-32 parallel-jaw gripper, two GelSlim sensors~\cite{taylor2021iros}, one Mitsubishi F/T sensor $1$F-FS$001$-W$200$ (only used to train the F/T RL policy training), four training objects and their resetting fixtures, and four environments for insertion, including line walls, corner walls, U walls and holes. 
%This is to prevent the learned policy from overfitting to one specific moving direction.
Note that the fixtures on the sides of the experimental setup are used for automatic object pose resetting for the training objects. Note also that the clearance of the U wall and the hole environments is 3mm, while the average width of the objects is 35mm.  

\subsection{Insertion Experiment}
Similar to~\cite{dong2019tactile}, we assume that during actual operation, a noisy position of the hole would be estimated by a vision system, and the objective of the tactile-RL policy is to correct for the remaining relative pose error. To emulate this scenario, we introduce random translation error ranging from -6 $\sim$ 6 mm in x and y axis (17\% of the object's width) and rotation error in yaw angle ($\theta$) ranging from \ang{-10} $\sim$ \ang{10}. The resetting mechanism for objects, allows to pick them with controlled noise, different grasping forces and different heights. This is helpful to avoid overfitting to the height of the object or certain local tactile image features.

\myparagraph{Experimental process} For every episode, the robot randomly chooses and grasps an object and moves it to the top of the insertion environment with additional translational and rotational error. Subsequently, the robot moves down vertically to attempt an insertion, while monitoring collisions with slip detection~\cite{dong2018slip}. If the slip detector is not triggered while moving down, that indicates a successful insertion. Otherwise, the insertion is blocked by the environment. We capture the tactile image sequences (or the F/T data when using the F/T policy) from the two tactile sensors during the contact and feed it to the controller. The robot moves the object to the next position according to the output of the controller, and starts another trial until either the object has been inserted or the insertion has failed. The robot resets the pose of the object by returning the object to its fixture after every $10$ trials during the training process.

\subsection{Deep RL Policy}
We choose an off-policy RL algorithm, the Twin Delayed DDPG (TD3)~\cite{TD3}, to train the tactile-RL policy for better sample-efficiency. The underlying RL framework is represented by a tuple of ($S, A, p, R, \gamma$). Here we use $12$ images (downsampled from 30 images within 0.5 seconds of contact to avoid the gradient vanishing problem in the RNN model, with $640 \times 480$ pixel resolution) captured by each tactile sensor (24 in total) during the contact period as the state $s_t \in S$. The terminal states are either the successful insertion state, or the state that has error larger than a threshold in any axis, or the state when the attempted number of insertions is over a limit. We set the max error $E_{max}$ in x and y axis to be 12mm, and \ang{15} in $\theta$. The maximum number of attempts is 15. The action $a_t \in A$ is the continuous robot displacement in $(\Delta x, \Delta y, \Delta \theta)$ in the current gripper frame. The reward function $R$, shown in (1), is calculated with the difference between the current contact error ($e_t$) and the previous contact error ($e_{t-1}$). A constant penalty term is given for each insertion attempt ($P$), and additional reward ($R_s$) is given to a successful insertion ($\alpha=1$, otherwise $\alpha=0$). 

\vspace{-7pt}
\begin{equation}
\vspace{-3pt}
    R_t = e_{t-1} - e_t - P + \alpha R_s
    % \vspace{-15pt}
\label{eq:reward}
\end{equation}

We use a discount factor $\gamma-0.99$ for computing the Q function. As mentioned in Sec.~\ref{sec:method}, we use curriculum learning for RL training, and we stop training when the mean of the rewards crosses a threshold and the standard deviation is lower than a threshold in the last $30$ episodes.

% Based on our observation, the tactile signal contains nearly no information when the error is over the limit, and it is too challenging for the policy to correct the error.

\myparagraph{Model architecture} TD3 is an actor-critic method~\cite{TD3}. We use a CNN+RNN+MLP model for the actor network with the 2 tactile image sequences used as input shown on the right-hand side of Fig.~\ref{fig:setup_NN}. Since the true state (the contact errors) of the task is known and thanks to precise initial grasps and careful resetting mechanisms during training, we directly use the current contact error in $(x, y, \theta)$ axis and the action $a_t$ from the actor model as the input to the critic model, instead of the tactile images. For the critic model, we use a small MLP model to estimate the $Q-$function. This can potentially speed up the convergence of the RL policy. %More details of the model architecture are provided in the supplementary materials.

\myparagraph{Training details} To further accelerate training of the RL policy, we bootstrap the actor network with a supervised learning policy trained with $300$ data points collected under a random policy in the first environment. In the first $50$ episodes of RL training, we freeze the actor network to avoid updates by the untrained critic network. This helps to accumulate positive examples into the replay buffer.% and guides both the actor and critic networks to converge quickly. 

\subsection{Supervised Learning Policy}
To compare with previous work~\cite{dong2019tactile}, we train a supervised learning (SL) policy with the data (tactile image sequence) and labels (object-hole misalignment) saved during the RL policy training as a baseline. The SL policy takes the marker flow image sequence as the input and outputs the estimated contact error in $x, y$ and $\theta$ axis. For a fair comparison, we use the same architecture for the SL policy as the RL actor model. The SL policy is trained to estimate the contact error with an MSE loss. During execution, we use the action which is the opposite of the output of the SL policy to minimize the contact error iteratively. The maximum number of attempts is also 15.

\subsection{RL policy with F/T sensor}
We train a RL policy with F/T sensing feedback as another baseline. We collect a stream of F/T signals ($32$ samples for each of the $6$ dimensions of F/T, downsampled from 120) during each contact (0.5 s) as the input to the RL policy. The downsampling is used to avoid the gradient vanishing problem for the RNN model, but still keep the original force profile. We remove the torque about the vertical axis $T_z$ from the input since it only change minimally during contact. We keep the general RL framework and only modify the neural network architecture of the actor network to a smaller RNN + MLP model to adapt to the data format of the F/T sensor. Since the model size is much smaller, we can train the policy directly in the hole environment with a SL policy initial bootstrap until the policy converges.

% data preprocessing 
% 

\section{Results}
\label{results}
We evaluate the performance of the polices by conducting 250 insertion experiments under different initial pose errors, with each of the 4 training objects and each of the 4 novel objects shown in Fig~\ref{fig:setup_NN}. The initial pose errors are uniformly sampled in the error space, where the errors in $x$ and $y$ axis range from -5mm to 5mm and the error in $\theta$ is from $-10^{\circ}$ to $10^{\circ}$. The performance metric includes the success rate and the average number of attempts, which is averaged only over the successful cases. 
\begin{table*}[t]
\centering
% \vspace{6pt}
\caption{Performance of different policies tested with 4 training objects and 4 novel objects } \label{tab:result} 
\begin{tabular}{l|p{1.0cm}|p{1.1cm}p{1.3cm}p{1.3cm}p{0.9cm}|p{0.85cm}p{0.9cm}p{0.9cm}p{0.8cm}}
\hline   %\vtop{\hbox{\strut Water}\hbox{\strut bottle}}
   &  &  Cylinder & Hexagonal cylinder &  Elliptical cylinder & Cuboid & Big bottle & Small bottle & Phone charger & Paper box \\\hline 
 \multirow{2}{*}{\parbox{0.8cm}{RL\mysubscript{*}}}& Success & 97.1\%  & \textbf{97.1 \%} & 98.0\% & \textbf{89.6\%} & 80.8\% & \textbf{96.7\%} & \textbf{96.8\%} & \textbf{70.1\%}  \\
 & Attempt & 2.96 & 3.83  & \textbf{2.34} & \textbf{5.42} & 3.53 & \textbf{2.10} & \textbf{2.55} & \textbf{7.91} \\ \hline 
 \multirow{2}{*}{\parbox{0.9cm}{SL}}& Success &  84.7\%  & 70.0\% & 93.7\% & 15.2\% & 27.0\% & 76.3\% & 42.46\% & 18.3\%  \\
 & Attempt & 3.04 & 3.34  & 2.60 & 3.83 & 4.23 & 2.02 & 5.10 & 3.58 \\ \hline 
 \multirow{2}{*}{RL w/o Curr}& Success & 99.1\%  & 89.5\% & 98.0\% & 57.1\% & 30.3\% & 36.23\% & 71.2\% & 9.01\%  \\
 & Attempt & 2.97 & 5.41  & 4.15 & 11.3&  8.79&  11.8&  6.69& 7.32  \\ \hline 
 \multirow{2}{*}{{RL RGB}}& Success & 98.8\%  & 93.1\% & \textbf{99.15\%} & 88.0\% & 37.8\% & 95.7\% & 74.1\% & 13.4\%  \\
 & Attempt & 2.26 & \textbf{2.65}  & 2.42 & \textbf{5.35} & 5.49 & 3.16 & 4.92 & 8.35 \\ \hline 
 \multirow{2}{*}{\parbox{0.9cm}{RL F/T}}& Success & \textbf{100.0}\%  & 89.3\% & \textbf{99.15\%} & 61.2\% & \textbf{99.5\%} & 95.2\% & 53.5\% & 54.1\%  \\
 & Attempt & \textbf{1.82} & 3.26  & 2.52 & 5.35 & \textbf{2.48} & 3.82 & 6.53 & 8.35 \\ \hline 
 \multirow{2}{*}{\parbox{0.9cm}{RL SRL}}& Success & \textbf{100.0}\%  & 89.3\% & 91.8\% & \textbf{98.0}\% & \textbf{99.5\%} & 95.2\% & 53.5\% & 54.1\%  \\
 & Attempt & \textbf{1.82} & 3.26  & 2.68 & \textbf{3.56} & \textbf{2.48} & 3.82 & 6.53 & 8.35 \\ 
 \hline 
\vspace{-17pt}
\end{tabular}
\end{table*}

%  \multirow{2}{*}{RL Abstract}& Success & \textbf{100.0}\%  & 89.3\% & \textbf{99.15\%} & \textbf{94.3\%} & \textbf{99.5\%} & 95.2\% & 53.5\% & \textbf{79.0\%}  \\
%  & Attempt & \textbf{1.82} & 3.26  & 2.52 & \textbf{4.17} & \textbf{2.48} & 3.82 & 6.53 & \textbf{6.10} \\ \hline 

% \begin{table*}[t]
% \centering
% \vspace{6pt}
% \caption{Performance of different policies tested with 4 training objects and 4 novel objects } \label{tab:result} 
% \begin{tabular}{l|l|l}
% \hline   %\vtop{\hbox{\strut Water}\hbox{\strut bottle}}
%  Policy  &  Training objects  & Novel objects  \\\hline 
% RL\mysubscript{*} & \textbf{95.4\%}  & \textbf{86.1}\%  \\ \hline 
% SL &  65.9\%   & 41.0\%   \\ \hline 
% RL w/o Curr & 85.9 & 36.7\%   \\ \hline 
% RL RGB & 94.6\%   & 55.2\%   \\ \hline 
% RL F/T & 87.4\% & 75.6\%     \\ \hline 

% \vspace{-10pt}
% \end{tabular}
% \end{table*}

\subsection{RL Policy with Curriculum and Tactile Flow (RL\mysubscript{*} policy)}
% how much needed to converge in each enviroment 
% the performance of the RL policy in both trained and novel objects
% convergence (maybe we show the learning curve)
\textbf{Convergence} of the RL\mysubscript{*} policy is surprisingly fast. Due to the warm start of the actor network, the policy can already insert the cylinder most of the time in the line-wall environment at the start. Since some object-specific features (texture and shape) of individual objects are removed in the marker flow, the policy trained with 1 object can quickly adapt to the other three objects. Because the policy from the first environment can already perform well in the corner-wall environment, we stop training after 25 episodes (50 data points, which means 2 attempts each episode). The U-shaped environment is much harder compared to previous 2 environments, and the policy needs around 150 episodes (500 data points) to adapt. The policy converges after 300 episodes (2000 data points) in the hole environment, with the warm start from the previous environment. From our observation, the policy can quickly learn how to insert the first three objects, but spends most of the time to learn how to correct the rotation error for the cuboid object. The whole training process uses around 8 hours, 3000 data in total. %More training details, such as learning curves, are shown in the supplementary materials. 
 
The \textbf{performance} of the RL\mysubscript{*} policy in the test experiments is illustrated in Table~\ref{tab:result}. For the 4 training objects, the RL\mysubscript{*} policy achieves 90\% success rate with the cuboid and over 97\% success rate for the remaining 3 objects. The cuboid object needs around $5\sim6$ attempts to be inserted, whereas the other three objects only need $2\sim3$ attempts. This result agrees with the respective difficulty of the tasks. As we described earlier, whereas the policy needs to correct the rotation error of cuboid to complete insertion, the other three objects are not sensitive to rotational error. Misalignment in rotation is hard to observe, especially when accompanied by errors in $x$ and $y$ axis. This is because the tactile signal is generated by a complex combination of object's rotation into and rotation parallel to the sensor surface. The policy can also generalize well to 4 novel objects. It achieves over 96\% success rate with the small bottle and the phone charger. Both of them are either round or have round corners, which makes the insertion easier. The success rate of the big bottle drops to 80\%, because the chamfer on the bottom edge changes the way the object rotates. The box can be inserted in over 70\% of the tests, which has the lowest success rate, but is also the most difficult task. Compared to the training object cuboid, the success rate is lower but the RL\mysubscript{*} policy still achieves good performance. The novel object and training objects requires similar number of attempts.

Most of the \textbf{failure} cases happen when both the translational (in $x$ and/or $y$) and rotational components (in $\theta$) of the initial pose error are large. Since the rotation of the object is too little to be captured by the tactile sensors in these cases, the RL policy can output an action in the opposite direction, resulting in a even bigger error, until it finally diverges. %This is a limitation of the approach and possible solutions include 1) adding vision to the system to help the localization~\cite{yu2018realtime, lee2018making}, 2) adding an informative exploration strategy around the contact point/line, to gather more information. 

\subsection{Baseline Comparison}
% We compare the RL\mysubscript{*} policy with the following three baseline policies: 1) the SL policy, 2) the RL policy training without curriculum, and 3) the RL policy with RGB images as input. 

\myparagraph{SL policy} According to Table~\ref{tab:result}, the SL policy performs a bit worse in the first 3 training objects, compared to the RL\mysubscript{*} policy, but can barely insert the cuboid during the test. In the test cases with 4 novel objects, only the small bottle is inserted with over 70\% success rate, and the others have significantly lower success rate. The RL\mysubscript{*} policy is also more efficient as to the number of attempts. We find that the cuboid objects are particularly challenging for the SL policy, due to the fact that the tactile signals are similar in many contact configurations, and it's very difficult for the SL policy to make the right move. In addition, considering the SL policy is trained with MSE loss, a safe choice to the controller is to output $0$ when the signal is not distinctive, which makes the SL policy get stuck. This comparison demonstrates that the RL policy learns a better strategy to insert objects even with some partially observable states. 

\myparagraph{RL policy without curriculum} To evaluate the improvement of the curriculum training on the convergence speed, we train another RL policy directly in the hole environment with all 4 training objects. To keep it consistent with RL\mysubscript{*} policy, we also bootstrap the RL policy with the same SL policy, and use maker flow as the input. We stop the training when it reaches the same amount of data used for RL\mysubscript{*} policy training. The performance of this policy is shown in the 4th row of Table~\ref{tab:result}. The policy achieves a similar success rate for the first three objects, but uses nearly twice the number of attempts compared to RL\mysubscript{*}. The policy fails in almost all of the cuboid insertions with large rotational error, and only gets 57\% success rate, using 11 attempts on an average. Based on the comparison, we can interpret that 1) localizing the contact direction in x and y axis is easy to learn, 2) optimizing the number of attempts, relating to precisely localizing the amplitude of the contact error and a high-level insertion strategy to quickly gather information, is harder, 3) estimating the rotational error is also difficult and data-intensive. This RL policy does not generalize to the new objects well, and only achieves 30\% average success rate. 

\myparagraph{RL policy with RGB input} The raw RGB tactile image contains task-relevant as well as detailed object-specific features. It is much easier for the RL policy using RGB images to overfit to the training objects. This hypothesis agrees well with the result shown in Table~\ref{tab:result}, 5th row. The performance of the policy with training objects is as good as that of the RL\mysubscript{*} policy. However, the success rate drops dramatically in the test of novel objects, especially with the big bottle and paper box. The policy keeps outputting the same action during the testing of these two objects, which reveals that the policy is likely overfitted. Using a proper tactile representation is important for generalization. Marker flow may not be the best representation, and we leave the problem of learning a good representation for future work. 

\myparagraph{RL policy with F/T sensing} achieves almost 100\% success rate with cylindrical objects with round edges in both the training (cylinder and elliptical cylinder) and novel objects (2 bottles), shown in Table~\ref{tab:result}, 6th row. In some of the objects, it uses even fewer number of attempts compared to RL* policy. However, its success rate drops to $89\%$ for hexagonal ellipse and further drops to around $55\%$ for the three cuboid shaped objects (cuboid, phone charger, paper box). According to our observation, the policy with F/T sensor is good at localizing contact errors in x and y directions, but has trouble to distinguish the rotation error. Compared to F/T sensor, the tactile sensor observes not only the force information, but also the object motion in the contact phase, which may contain useful information to differentiate the rotation error.

\section{Conclusion}
\label{sec:conclusion}
%===============================================================================
In this paper, we proposed a RL insertion policy with tactile feedback from 2 GelSlim~\cite{taylor2021iros} tactile sensors without prior knowledge of the object's geometry. We show that our proposed RL policy with designed curriculum training and tactile flow representation, provides several advantages over other tactile-based baseline policies, including: 1) supervised learning policy, 2) RL policy without curriculum, 3) RL policy with RGB input. Furthermore, we also showed the advantage of using tactile over F/T sensors in terms of the generalization, which are the most widely used sensors for these kind of applications.

\bibliographystyle{IEEEtran}
\bibliography{Ref}

\end{document}